\documentclass[twocolumn]{svjour3} 
\smartqed
\usepackage{natbib}
\usepackage{graphicx}
\usepackage{colortbl}
\usepackage[dvipsnames]{xcolor}
\usepackage{booktabs,multirow,array}
\definecolor{lightgray}{gray}{0.9}
\definecolor{cvprblue}{rgb}{0.21,0.49,0.74}
\usepackage[pagebackref,breaklinks,colorlinks,bookmarks=false,citecolor=cvprblue]{hyperref}
\usepackage{ragged2e}
\usepackage[ruled]{algorithm2e}
\usepackage{algorithmic}
\usepackage[misc]{ifsym}

\usepackage{amsmath}
\usepackage{amsfonts}       %
\usepackage{algorithmic}
\usepackage[caption=false,font=footnotesize]{subfig}
\usepackage{ragged2e}
\usepackage{enumitem} %

\usepackage{wrapfig,lipsum}
\usepackage{caption}
\usepackage{subcaption,siunitx}
\let\titleold\title
\renewcommand{\title}[1]{\titleold{#1}\newcommand{\thetitle}{#1}}

\newcommand{\ie}{\emph{i.e.}}
\newcommand{\eg}{\emph{e.g.}}
\newcommand{\vs}{\emph{vs.}}

\usepackage{amsmath,amsfonts,bm}

\def\Secref#1{Section~\ref{#1}}

\def\eqref#1{equation~\ref{#1}}
\def\Eqref#1{Equation~\ref{#1}}

\def\1{\bm{1}}

\def\vx{{\bm{x}}}

\def\vz{{\bm{z}}}

\def\mE{{\bm{E}}}

\def\mK{{\bm{K}}}

\def\mP{{\bm{P}}}
\def\mQ{{\bm{Q}}}
\def\mR{{\bm{R}}}

\def\mV{{\bm{V}}}
\def\mW{{\bm{W}}}
\def\mX{{\bm{X}}}

\def\mZ{{\bm{Z}}}

\DeclareMathAlphabet{\mathsfit}{\encodingdefault}{\sfdefault}{m}{sl}
\SetMathAlphabet{\mathsfit}{bold}{\encodingdefault}{\sfdefault}{bx}{n}

\begin{document}

\title{Collect-and-Distribute Transformer for 3D Point Cloud Analysis}

\author{
Haibo Qiu $\cdot$ Baosheng Yu $\cdot$  Dacheng Tao
}

\institute{
Haibo Qiu $\cdot$ Baosheng Yu $\cdot$  Dacheng Tao \\
\texttt{\{hqiu2518, baosheng.yu, dacheng.tao\}@sydney.edu.au} \\
\\
School of Computer Science, the University of Sydney, Sydney NSW, 2008, Australia 
}

\date{Received: date / Accepted: date}
\maketitle

\begin{abstract}
Remarkable advancements have been made recently in point cloud analysis through the exploration of transformer architecture, but it remains challenging to effectively learn local and global structures within point clouds. In this paper, we propose a new transformer network equipped with a collect-and-distribute mechanism to communicate short- and long-range contexts of point clouds, which we refer to as CDFormer. Specifically, we first employ self-attention to capture short-range interactions within each local patch, and the updated local features are then collected into a set of proxy reference points from which we can extract long-range contexts. Afterward, we distribute the learned long-range contexts back to local points via cross-attention. To address the position clues for short- and long-range contexts, we additionally introduce the context-aware position encoding to facilitate position-aware communications between points. We perform experiments on five popular point cloud datasets, namely ModelNet40, ScanObjectNN, ShapeNetPart, S3DIS and ScanNetV2, for classification and segmentation. Results show the effectiveness of the proposed CDFormer, delivering several new state-of-the-art performances on point cloud classification and segmentation tasks. The source code is available at \url{https://github.com/haibo-qiu/CDFormer}.
\end{abstract}
\section{Introduction}
\label{sec:intro}

Point clouds have been extensively investigated in recent years, mainly due to their numerous promising real-world applications, such as autonomous driving~\citep{aksoy2020salsanet,li2020deep} and robotics~\citep{li2019integrate, YANG2020101929}. Unlike 2D images, a point cloud is a set of 3D points distributed in an irregular and disordered manner, where each point is usually characterized by its Cartesian coordinates $(x, y, z)$. These fundamental differences make it non-trivial to utilize off-the-shelf 2D deep architectures for point cloud analysis. Therefore, many recent methods have developed different deep architectures to handle the special characteristics of point clouds, including mlp-based~\citep{ma2022rethinking,qi2017pointnet,qi2017pointnet++}, cnn-based~\citep{li2018pointcnn}, and graph-based models~\citep{thomas2019kpconv,wang2019dynamic}. These architectures aim to address the challenges posed by the irregularity and unordered nature of point clouds and enable effective point cloud analysis.

One of the main challenges in point cloud analysis is efficiently exploring local and global features while considering the irregular and disordered characteristics of point clouds. Recently, transformer architectures, which enable effective local/global-range learning via the attention mechanism, have become popular in both natural language processing~\citep{brown2020language,devlin2018bert,vaswani2017attention} and computer vision~\citep{dosovitskiy2020image,liu2021swin,wang2021pyramid,wang2022pvt}. Inspired by this, transformer architectures have been further introduced for point cloud analysis~\citep{guo2021pct,lai2022stratified,park2021fast,zhao2021point}. However, the vanilla self-attention module in transformer has a time complexity of $\mathcal{O}(N^2)$ when operating on a sequence of tokens with length $N$. When taking each point as a token, the $\mathcal{O}(N^2)$ complexity becomes unaffordable, as there are tens of thousands of points for each point cloud in real-world applications~\citep{armeni20163d}. To address this challenge, \cite{zhao2021point} introduces vector self-attention in a local way, avoiding $\mathcal{O}(N^2)$ complexity by only interacting features with $K$ neighbors (\eg, $K=16$), while failing to capture long-range contexts. On the other hand, \cite{lai2022stratified} employs local window-based self-attention with a shifted window strategy similar to~\cite{liu2021swin} and proposes to capture long-range contexts by sampling nearby points densely and distant points sparsely. Nevertheless, due to the heterogeneous density distribution of point clouds, the fixed-size window partition employed in~\cite{lai2022stratified} leads to a diverse number of points in each local window, thus requiring complicated and sophisticated designs during implementation.

\begin{figure}
   \centering
   \includegraphics[width=.9\linewidth]{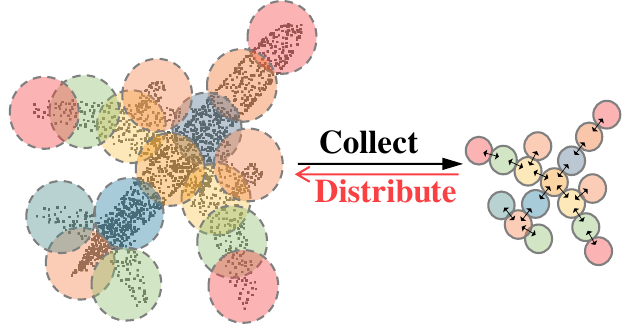}
   \caption{An illustration of the proposed collect-and-distribute mechanism.}
   \label{fig:cd}
\end{figure}

In this work, we propose a new collect-and-distribute transformer, CDFormer, to learn both short- and long-range contexts for 3D point cloud analysis. Specifically, we first divide the point cloud into a set of local patches using $K$ nearest neighbor points, instead of a fixed window partition~\citep{lai2022stratified}. Each local patch contains the same number of points, enabling direct modeling by popular deep learning packages~\citep{abadi2016tensorflow,paszke2019pytorch} without custom operations and avoids the prohibitive $\mathcal{O}(N^2)$ time complexity. Besides local self-attention for short-range interactions within each local patch, we introduce a collect-and-distribute mechanism. This mechanism first collects local patch information to a set of proxy reference points, explores long-range contexts among these reference points, and distributes the collected long-range contexts back to local points through cross-attention between reference points and local points. An illustration of the proposed collect-and-distribute mechanism is shown in Figure~\ref{fig:cd}. To enhance local-global structure learning, the positional information is critical for transformers employed in point clouds~\citep{lai2022stratified,zhao2021point}. Therefore, we introduce context-aware position encoding for CDFormer, where the relative position information interacts with the input features to dynamically enhance the positional clues.

\noindent
Our main contributions can be summarized as follows:
\begin{itemize}
    \item We propose CDFormer to capture short- and long-range contexts simultaneously with a novel collect-and-distribute mechanism, effectively learning local-global structures.
    \item We introduce context-aware position encoding to enhance position clues and facilitate communications within points.
    \item We perform extensive experiments on five well-known point cloud datasets: ModelNet40~\citep{wu20153d} and ScanObjectNN~\citep{uy2019revisiting} for classification, along with ShapeNetPart~\citep{yi2016shapnetpart}, S3DIS~\citep{armeni20163d}, and ScanNetV2~\citep{dai2017scannet} for segmentation. The experimental results and analysis demonstrate the effectiveness of CDFormer and achieve state-of-the-art performances in point cloud processing.
\end{itemize}
\section{Related Work}
\label{sec:related}

\textbf{Transformer Architectures.}
Transformer architectures have emerged as a dominant framework for natural language processing in recent years~\citep{devlin2018bert, vaswani2017attention}. In addition, they have seen widespread exploration in the realm of vision tasks, with ViT~\citep{dosovitskiy2020image} being a groundbreaking work that divides images into local patches and treats each patch as a token. Building upon the success of ViT, numerous subsequent works have been proposed that either explore hierarchical architectures with multi-scale resolutions~\citep{fan2021multiscale, gu2022multi, li2021improved, liu2021swin, wang2021pyramid, wang2022pvt} or incorporate local-global information~\citep{chu2021twins, li2022sepvit, yang2021focal}. For instance, PVT~\citep{wang2021pyramid} devises a progressive shrinking pyramid to effectively explore multi-resolution features while HRViT~\citep{gu2022multi} integrates high-resolution multi-branch architectures to learn multiplicative scale representations. Twins-SVT~\citep{chu2021twins} incorporates locally-grouped self-attention and global sub-sampled attention to capture local-global contexts. Swin Transformer~\citep{liu2021swin} introduces a hierarchical transformer architecture equipped with the shifted window strategy to enable cross-window communications. SepViT~\citep{li2022sepvit} proposes a self-attention mechanism that is depthwise separable to facilitate efficient local and global information exchange within a single attention block.

\noindent
\textbf{Point Cloud Analysis.} The mainstream point cloud analysis methods can be roughly divided into three directories: point-based~\citep{guo2021pct,lai2022stratified,li2018pointcnn,ma2022rethinking,park2021fast,qian2022pointnext,qi2017pointnet,qi2017pointnet++,qiu2023pointhr,tang2023erda,thomas2019kpconv,wang2019dynamic,xu2021you,zhao2021point}, voxel-based~\citep{cheng20212,tang2020searching,zhu2021cylindrical}, and projection-based~\citep{cortinhal2020salsanext,qiu2022gfnet,milioto2019rangenet++,zhang2020polarnet}. 
For a better trade-off between complexity and efficiency, we focus primarily on point-based approaches, where existing methods usually devise novel operations/architectures for raw points, including mlp-based~\citep{ma2022rethinking,qian2022pointnext,qi2017pointnet,qi2017pointnet++}, cnn-based~\citep{li2018pointcnn}, graph-based~\citep{thomas2019kpconv,wang2019dynamic}, and transformer-based~\citep{guo2021pct,lai2022stratified,park2021fast,zhao2021point}. A pioneering work in this field is PointNet~\citep{qi2017pointnet}, which directly processes point clouds using multi-layer perceptrons (MLPs). This approach was further improved upon by PointNet++\citep{qi2017pointnet++}, which introduced a hierarchical structure for processing point clouds. PointNext\citep{qian2022pointnext} proposes even more improved training strategies that significantly improve upon PointNet++\citep{qi2017pointnet++}. PointCNN\citep{li2018pointcnn} learns an x-transformation from the input points for alignment, which is followed by typical convolution layers. In contrast, KPConv~\citep{thomas2019kpconv} introduces kernel point convolution, a new point convolution operator, that takes neighboring points as input and processes them with spatially located weights. Recently, transformer architectures have been introduced for point cloud analysis~\citep{guo2021pct, lai2022stratified, park2021fast, zhao2021point}. PCT~\citep{guo2021pct} presents the offset-attention mechanism, which replaces the original self-attention. Point Transformer~\citep{zhao2021point} proposes to introduce a vector self-attention mechanism to aggregate neighbor features but fails to capture long-range dependencies. Stratified Transformer~\citep{lai2022stratified} captures long-range contexts by sampling nearby points densely and distant points sparsely with the shifted window strategy, as used in Swin Transformer~\citep{liu2021swin}. However, due to the varying density distribution of point clouds, partitioning windows in a fixed size leads to varying point counts in different local windows, requiring sophisticated designs to address this issue.

\section{Method}
\label{sec:method}

In this section, we first provide an overview of the proposed CDFormer for 3D point cloud analysis. We then introduce the patch division and the collect-and-distribute mechanism in detail. Lastly, we discuss the proposed context-aware position encoding.

\subsection{Overview}
A 3D point cloud usually consists of a set of $N$ points $\mP \in \mathbb{R}^{N \times 3}$, where each point is featured by the Cartesian coordinate $(x, y, z)$. The objective of typical point cloud analysis tasks is to assign semantic labels to the point cloud. For example, for point cloud classification, the goal is to predict a single semantic label $Q \in \{0, \cdots, U-1\}$ for the whole point cloud $\mP$, where $U$ is the total number of semantic categories. For point cloud segmentation, the goal is to assign a semantic label for each point in the point cloud $\mP$.

\begin{figure*}[!ht]
  \centering
  \includegraphics[width=\linewidth]{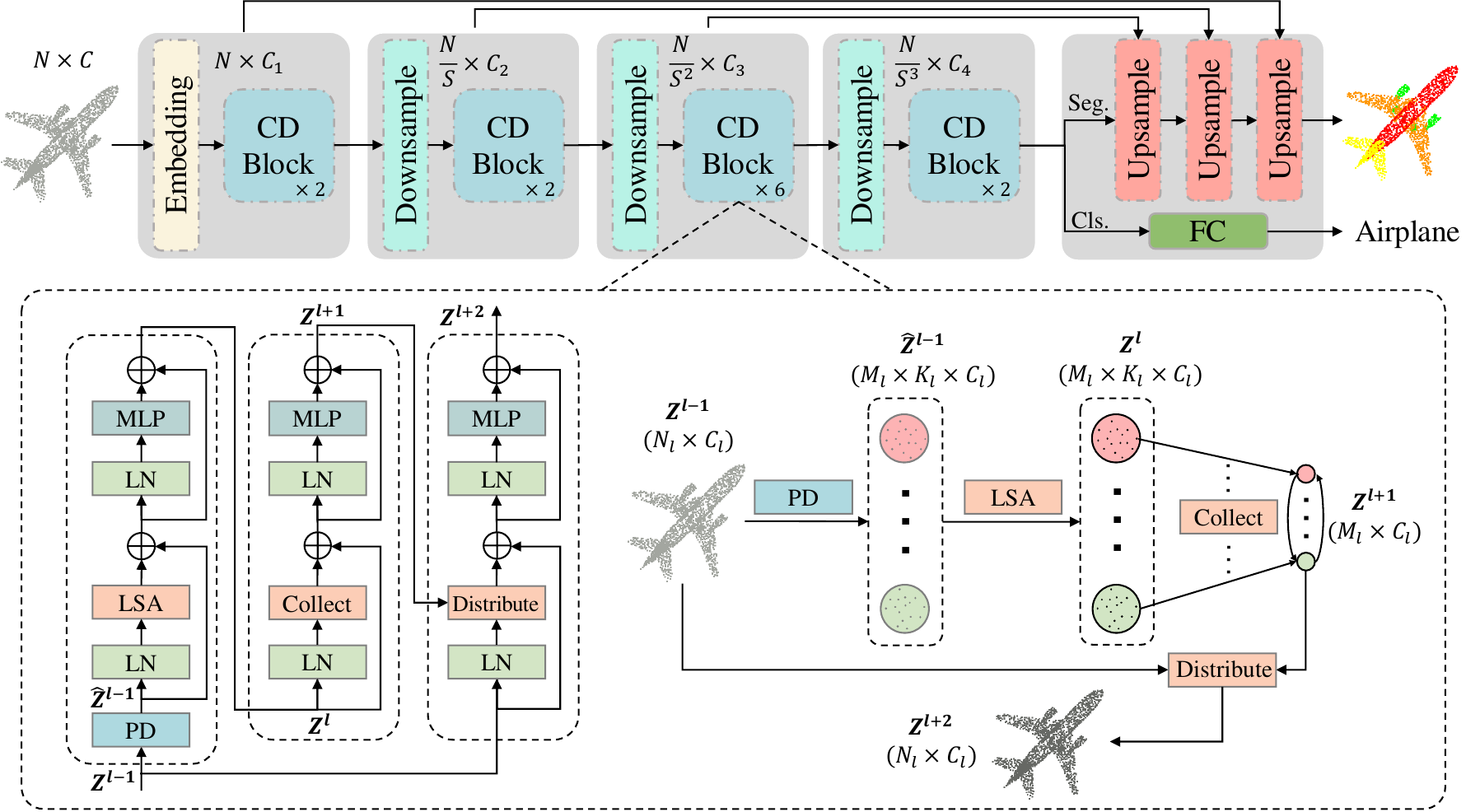}
   \caption{The main collect-and-distribute transformer framework. Note that \textbf{PD} refers to patch division and \textbf{LSA} indicates the local self-attention defined by~\Eqref{eq:lsa1} and~\ref{eq:lsa}.}
   \label{fig:framework}
\end{figure*}

The main CDFormer framework for 3D point cloud analysis is illustrated by Figure~\ref{fig:framework}. Since a point cloud may also contain extra features such as color, we consider the input feature of a general point cloud as $\mX \in \mathbb{R}^{N \times C}$, where $C$ indicates the number of feature channels. Specifically, we first utilize a KPConv~\citep{thomas2019kpconv} layer as the embedding layer to aggregate local information for raw point embeddings to obtain the embedded features with the size of $N \times C_1$. After that, the main backbone network is a stack of multiple the proposed collect-and-distribute blocks or CD Blocks, where each block first divides all points into local patches and then  explores short- and long-range contexts in a collect-and-distribute manner as follows: 1) a local self-attention is first used to learn short-range relations in each patch of points; 2) the learned local features are then collected to a set of proxy reference points which then communicate with each other to capture long-range contexts; and 3) the learned long-range contexts are finally distributed back to the original local patch points such that the learned point embeddings are equipped with both short- and long-range information. CDFormer is built by stacking multiple downsampling layers and CD Blocks.

\subsection{Patch Division}
\label{sec:pl}

For a typical point cloud with tens of thousands of points~\citep{armeni20163d}, it is computationally prohibitive to consider each point as a token: given a sequence of tokens with the length $N$, the time complexity of the self-attention in transformer is $\mathcal{O}(N^2)$. 
Therefore, following~\citep{dosovitskiy2020image,pang2022masked,yu2021point}, we divide a point cloud into multiple local patches, \eg, $M$ patches with $K$ points in each patch, and then employ self-attention within each local patch instead of all points. Therefore, with a proper patch division, it becomes acceptable with a linear time complexity $\mathcal{O}(MK^2)$.

We describe the patch division process used in our method as follows. Given a scale factor $S$, the furthest point sampling algorithm (FPS)~\citep{eldar1997farthest,qi2017pointnet++} is applied on the point features $\mX \in \mathbb{R}^{N \times C}$ to obtain patch centers $\bar{\mX} \in \mathbb{R}^{M \times C}$ where $M=N/S$. For each local patch, we group the $K$ nearest neighbors around each patch center and reformulate $M$ patches as $\hat{\mX} \in \mathbb{R}^{M \times K \times C}$, and then apply the local self-attention (LSA) to extract local features. Specifically, for the $i_{th}$ patch where $i \in [0, \cdots, M-1]$, let $\hat{\vx} \in \mathbb{R}^{K \times C}$ denote its representation, which is then fed into MLPs to generate the query token $\mQ_{lsa}=\mW^q_{lsa} \hat{\vx}$, the key token $\mK_{lsa}=\mW^k_{lsa} \hat{\vx}$, and the value token $\mV_{lsa}=\mW^v_{lsa} \hat{\vx}$, respectively. 
The output $\bar{\vz} \in \mathbb{R}^{K \times C}$ can thus be calculated as follows:
\begin{align}
\qquad \qquad \qquad \quad \bar{\vz} &= \mW^{attn}_{lsa} \mV_{lsa}, \label{eq:lsa1}\\ 
\mW^{attn}_{lsa} &= \text{Softmax}((\mQ_{lsa} \mK_{lsa}^{\top})/{\sqrt{C}}). \label{eq:lsa}
\end{align} 
All $M$ patches features can be similarly obtained as $\bar{\mZ} \in \mathbb{R}^{M \times K \times C}$. Notably, we do not discuss the position encoding here and leave it in Sec~\ref{sec:position}.
Lastly, we have the complexity as follows: the local self-attention in each patch only has the time complexity $\mathcal{O}(K^2)$, and the overall time complexity of $M$ patches become affordable $\mathcal{O}(MK^2)=\mathcal{O}(NK^2/S)$. Through LSA, all points in each local patch can communicate with each other to capture short-range dependencies.

\subsection{Collect-and-Distribute Mechanism}
\label{sec:cd}

\begin{figure*}[t]
    \centering
    \includegraphics[width=.8\linewidth]{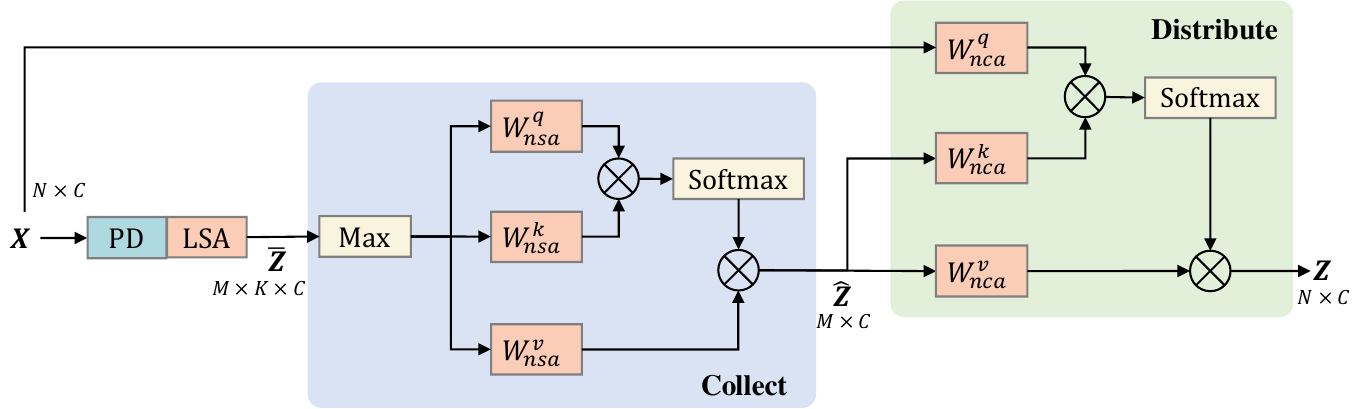}
    \caption{Collect-and-distribute block.  Note that \textbf{Collect} involves the neighbor self-attention (nsa) defined by~\Eqref{eq:nsa1} and~\ref{eq:nsa}; \textbf{Distribute} includes the neighbor cross-attention (nca) formulated in~\Eqref{eq:nca1} and~\ref{eq:nca}.}
    \label{fig:cdblock}
\end{figure*}

As the aforementioned local self attention only models the short-range information in each local patch, we then introduce how to explore long-range contexts with the proposed collect-and-distribute mechanism.
Specifically, we first collect those communicated local feature as a set of proxy reference points such that each local patch directly corresponds to a specific proxy point. To capture the long-range dependencies, a neighbor self-attention (NSA) is then applied on those proxy references to allow feature propagation among neighbors. By doing this, we can extract long-range contexts with a reduced linear time complexity since each proxy represents a patch of points. Lastly, those enhanced proxies distribute back to the local points via cross-attention to achieve short- and long-range contexts communications. An illustration of the proposed block is shown in Figure~\ref{fig:cdblock}, and we depict each step in detail as follows.

\begin{figure}[t]
    \centering
    \includegraphics[width=\linewidth]{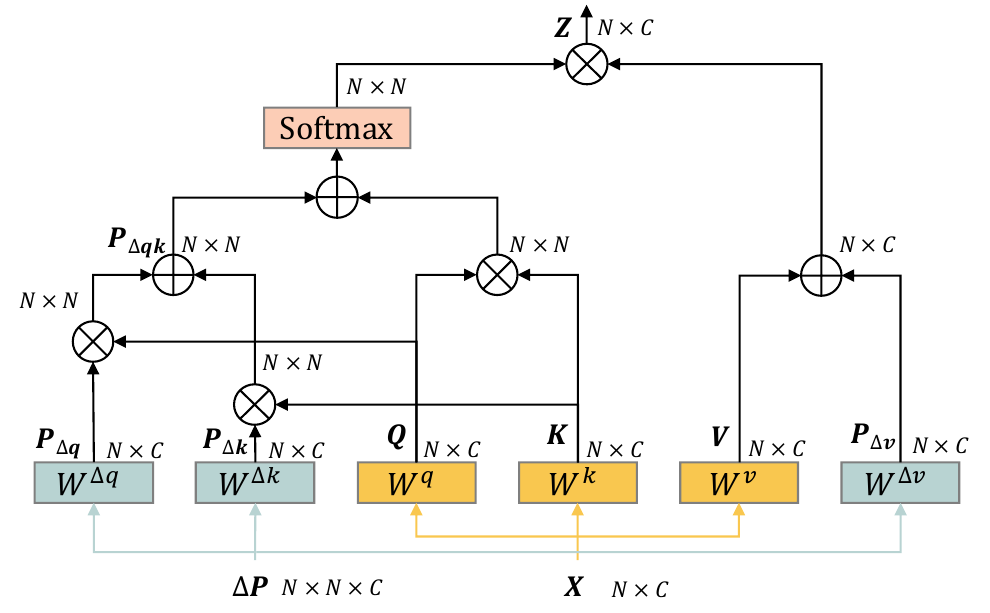}
    \caption{Context-aware position encoding.}
    \label{fig:position}
\end{figure}

\textbf{Collect}. 
Given local patches $\bar{\mZ} \in \mathbb{R}^{M \times K \times C}$, we collect local patch information from all $K$ local points to a proxy reference point via a max-pooling operation, $\mR =\text{maxpool}(\bar{\mZ}) \in \mathbb{R}^{M \times C}$. Next, we consider the $K$ nearest neighbors for each proxy $\hat{\mR} \in \mathbb{R}^{M \times K \times C}$ to capture the long-range contexts by employing a neighboring self-attention between $\mR$ and $\hat{\mR}$, \ie, a self-attention between each proxy and its $K$ neighbors followed by a sum operation to generate the output $\hat{\mZ} \in \mathbb{R}^{M \times C}$. We first generate the feature embeddings with MLPs for query, key, and value by $\mQ_{nsa} = \mW^q_{nsa} \mR$, $\mK_{nsa} = \mW^k_{nsa} \hat{\mR}$, and $ \mV_{nsa} = \mW^v_{nsa} \hat{\mR},$ respectively. Then we can formulate the process as follows: 
\begin{align}
\qquad \qquad \qquad \quad \hat{\mZ} &= \text{Sum}(\mW_{nsa}^{attn} \mV_{nsa}) \label{eq:nsa1}\\ 
\mW_{nsa}^{attn} &= \text{Softmax}((\mQ_{nsa} \mK_{nsa}^{\top})/{\sqrt{C}}).
\label{eq:nsa}
\end{align}
Notably, the time complexity $\mathcal{O}(NK^2/S)$ is linear to $N$ because the NSA is only applied on $K$ neighbors instead of all proxies.

\textbf{Distribute}. We distribute the long-range information in $\hat{\mZ} \in \mathbb{R}^{M \times C}$ back to local patch points for joint short- and long-range contexts via a neighbor cross-attention (NCA). In particular, given vanilla point features $\mX \in \mathbb{R}^{N \times C}$ and enhanced proxies $\hat{\mZ} \in \mathbb{R}^{M \times C}$, we first group the $K$ nearest neighbor proxies $\mE \in \mathbb{R}^{N \times K \times C}$ for each point. After that, we employ the neighbor cross-attention by regarding $\mX$ as the query, $\mE$ as the key and value, which is fed into a successive sum operation to obtain the final output $\mZ \in \mathbb{R}^{N \times C}$. We first get the embeddings with MLPs by $\mQ_{nca} = \mW^q_{nca} \mX$, $\mK_{nca} = \mW^k_{nca} \mE$, and $ \mV_{nca} = \mW^v_{nca} \mE$, respectively. After that, we can calculate $\mZ$ as follows:
\begin{align}
\qquad \qquad \qquad \quad \mZ &= \text{Sum}(\mW_{nca}^{attn} \mV_{nca}) \label{eq:nca1}\\
\mW_{nca}^{attn} &= \text{Softmax}((\mQ_{nca} \mK_{nca}^{\top})/{\sqrt{C}}).
\label{eq:nca}
\end{align}
In this way, each point communicates with $K$ enhanced proxies, \ie, approximate $K^2$ original proxies, and $K^3$ original points. Therefore, the long-range dependencies can be effectively distributed back to local points, \ie, both short- and long-range contexts are fused into the learned representations.

\subsection{Context-Aware Position Encoding}
\label{sec:position}

The position information of input tokens is necessary for transformer architectures~\citep{dosovitskiy2020image,vaswani2017attention}, while 
3D point cloud naturally contains the $(x,y,z)$ coordinates as position. 
Therefore, it might be straightforward to remove position encoding in the transformer-based architecture for point clouds. However, when the network goes deeper, the coordinates information can not keep precisely intact~\citep{lai2022stratified,zhao2021point}. Inspired by~\cite{wu2021rethinking}, we propose to further enhance the position clues in point cloud transformer via a  context-aware position encoding (CAPE). Specifically, CAPE is calculated by interacting relative position differences with the current features, thus can simultaneously handle the unordered characteristic of the point cloud and adaptively enhance the position information. An illustration of the proposed context-aware position encoding is shown in Figure~\ref{fig:position}. 

Given the input $\mX \in \mathbb{R}^{N \times C}$ and the relative position differences $\Delta \mP \in \mathbb{R}^{N \times N \times 3}$, we first obtain the $\mQ, \mK, \mV$ and their position embeddings $\mP_{\Delta q}, \mP_{\Delta k}, \mP_{\Delta v}$ via MLPs. We then calculate the CAPE by:
\begin{equation}
 \qquad \qquad \quad   \mP_{\Delta qk}=\mP_{\Delta q} \mQ^{\top} + \mP_{\Delta k} \mK^{\top},
\end{equation}
such that it is aware of the input features and can dynamically magnify the position information to effectively facilitate the communication of points. Lastly, we obtain the output $\mZ \in \mathbb{R}^{N \times C}$ as follows:
\begin{align}
\qquad \qquad \quad \mZ &= \mW_{attn}\cdot(\mV + \mP_{\Delta v}) \\ 
\mW_{attn} &= \text{Softmax}((\mQ \mK^{\top} + \mP_{\Delta qk})/{\sqrt{C}}).
\label{eq:cape}
\end{align}
Notably, if not otherwise stated, we use CAPE in all attention layers, including LSA, NSA, and NCA, to better capture short- and long-range contexts.

\section{Experiments}
\label{sec:experiment}

In this section, we first introduce the implementation details. We then evaluate the proposed CDFormer on ModelNet40~\citep{wu20153d} and ScanObjectNN~\citep{uy2019revisiting} for point cloud classification, ShapeNetPart~\citep{yi2016shapnetpart} for point cloud part segmentation, as well as S3DIS~\citep{armeni20163d} and ScanNetV2~\citep{dai2017scannet} for point cloud scene segmentation. Lastly, we perform comprehensive ablation studies on each component.

\subsection{Implementation Details.}
\label{sec:setup}

For ModelNet40~\citep{wu20153d}, following~\cite{qian2022pointnext}, we use fewer blocks in each stage $[1,1,3,1]$ with larger feature dimensions $\mathcal{C}=[64,128,256,512]$ and larger number of heads $\mathcal{H}=[4,8,16,32]$. We simply use $K=16$ neighbours in all blocks. Cosine annealing schedule with learning rate 0.001 is adopted for training 600 epochs with batch size 32. We use AdamW optimizer~\citep{loshchilov2017decoupled} with the weight decay 0.05. For each data sample, only 1,024 points with $(x, y, z)$ are used for training and testing, and common data augmentations, including shift, scale and cutmix~\citep{zhang2022pointcutmix}, are employed. The downsampling scale $S$ of each stage is set to 4. 
For ScanObjectNN~\citep{uy2019revisiting}, we follow~\cite{qian2022pointnext} to use point resampling to adapt 1,024 points for training, and only the hardest perturbed variant (PB\_T50\_RS) is considered in our experiment. We keep other training configurations the same as those in ModelNet40.

For S3DIS~\citep{armeni20163d}, following the practice in~\cite{lai2022stratified,zhao2021point}, we first apply the grid sampling on the raw input points with the grid size $0.04$m. We adopt the encoder with four stages with $[2, 2, 6, 2]$ blocks, the number of channels $\mathcal{C} = [C_1, C_2, C_3, C_4] = [48,96,192,384]$ and the number of heads $\mathcal{H} = [H_1, H_2, H_3, H_4] = [3,6,12,24]$. For simplicity, we set neighbours $K=16$ in all blocks. During the training process, the maximum number of input points is set to 80,000. Meanwhile, we set the downsampling scale $S$ of each stage to 8. We use AdamW optimizer~\citep{loshchilov2017decoupled} with the weight decay 0.01. All models are trained for 100 epochs, and the learning rate starts from 0.01 and drops by 1/10 at 60 and 80 epochs. We use four V100 GPUs with the batch size of 8. Following~\cite{qian2022pointnext}, we use the cross-entropy loss with label smoothing, and popular data augmentations including jitter, scale, rotate, and color drop.
For ScanNetV2~\citep{dai2017scannet}, we use grid sampling with a size of $0.02$m on the raw point clouds, following~\cite{wu2022point}. We employ the OneCycleLR~\citep{smith2019super} scheduler, which increases the learning rate from 0.0005 to 0.005 in the first 5 epochs and then uses cosine annealing to decrease it to 0 over the remaining 95 epochs. Additionally, we repeat the training set (1,201 scenes) 9 times to obtain 10,809 samples during the training process. 
For ShapeNetPart~\citep{yi2016shapnetpart}, we follow~\cite{qian2022pointnext} to use 2,048 points for training and testing. Considering that each point cloud has fewer points than it in S3DIS~\citep{armeni20163d}, we reduce the downsampling scale $S$ to 4 and use the batch size 80. The model is trained for 300 epochs, and the learning rate starts from 0.01 and drops by 1/10 at  210 and 270 epochs. Other settings are same as those in S3DIS.

\subsection{Point Cloud Classification}
\textbf{Dataset and Metrics.} ModelNet40~\citep{wu20153d} is a canonical dataset for object shape classification, which consists of 9,843 training and 2,468 testing CAD models belonging to 40 categories. We report the results of overall accuracy (OA). ScanObjectNN~\citep{uy2019revisiting} is a more challenging real-world benchmark in terms of background, noise, and occlusions. It contains totally 15,000 objects from 15 classes. We report the results of the mean of class-wise accuracy (mAcc) and overall accuracy (OA) on hardest perturbed variant, \ie, PB\_T50\_RS.
\begin{table}[t]
    \centering
    \caption{Results on ModelNet40.}
    \begin{tabular}{lcc}
    \toprule
         Method  & \#Points & OA (\%)    \\
         \midrule
         PointNet~\citep{qi2017pointnet}  & 1024 &89.2  \\
         PointNet++~\citep{qi2017pointnet++}  & 1024 &90.7   \\
         PointNet++~\citep{qi2017pointnet++} & 5000 &91.9  \\
         PointCNN~\citep{li2018pointcnn}   &1024&92.5 \\
         
         PointConv~\citep{wu2019pointconv} & 1024&92.5  \\
         A-CNN~\citep{komarichev2019cnn}    & 1024&92.6 \\
         KPConv~\citep{thomas2019kpconv}     & 7000 &92.9  \\
         DGCNN~\citep{wang2019dynamic}      &1024&92.9 \\
         PointASNL~\citep{yan2020pointasnl}  &1024&92.9 \\
         PointNext~\citep{qian2022pointnext} &1024&93.2 \\
         PosPool~\citep{liu2020closer}    &5000&93.2 \\
         PCT~\citep{guo2021pct}        &1024&93.2 \\
         SO-Net~\citep{li2018so}    &5000&93.4  \\
         PT~\citep{zhao2021point}  &1024&93.7  \\
         GBNet~\citep{qiu2021geometric} &1024 &93.8 \\
         PA-DGC~\citep{xu2021paconv}  &1024&93.9 \\
         PointMLP~\citep{ma2022rethinking} &1024 &\textbf{94.1} \\ %
         \midrule
         CDFormer \small{(ours)} &1024 &\underline{94.0} \\ %
         \bottomrule
    \end{tabular}
     \label{tab:cls_modelnet40}
\end{table}

\begin{table}[t]
    \centering
    \caption{Results on ScanObjectNN.}
    \begin{tabular}{lcc}
        \toprule
         Method & mAcc (\%) &OA (\%) \\
         \midrule
         PointNet~\citep{qi2017pointnet} &63.4 &68.2 \\
         SpiderCNN~\citep{xu2018spidercnn} &69.8 &73.7 \\
         PointNet++~\citep{qi2017pointnet++} &75.4 &77.9 \\
         DGCNN~\citep{wang2019dynamic} &73.6 &78.1 \\
         PointCNN~\citep{li2018pointcnn} &75.1 &78.5 \\
         BGA-DGCNN~\citep{uy2019revisiting} &75.7 &79.7 \\
         BGA-PN++~\citep{uy2019revisiting} &77.5 &80.2 \\
         DRNet~\citep{qiu2021dense} &78.0 &80.3 \\
         GBNet~\citep{qiu2021geometric} &77.8 &80.5 \\
         SimpleView~\citep{goyal2021revisiting} & - &80.5$\pm$0.3 \\
         PRANet~\citep{cheng2021net} & 79.1&82.1 \\
         MVTN~\citep{hamdi2021mvtn} & -&82.8 \\
         PointMLP~\citep{ma2022rethinking} & 83.9$\pm$0.5 & 85.4$\pm$0.3\\
         PointNext~\citep{qian2022pointnext} & 85.8$\pm$0.6 & 87.7$\pm$0.4 \\
         \midrule
         CDFormer \small{(ours)} &\textbf{87.2$\pm$0.3} &\textbf{88.4$\pm$0.2}\\
         
         \bottomrule
    \end{tabular} 
    \label{tab:cls_scanobjectNN}
\end{table}

\vspace{1mm}
\noindent
\textbf{Results.} Table~\ref{tab:cls_modelnet40} shows the results on ModelNet40, where the proposed CDFormer achieves comparable performance $94.0\%$ with other state-of-the-arts by only taking 1,024 points as input. In Table~\ref{tab:cls_scanobjectNN}, we also evaluate the proposed method on the most challenging variant (PB\_T50\_RS) of ScanObjectNN, where CDFomer achieves $87.2\pm0.3$ and $88.4\pm0.2$ on mAcc and OA, outperforming all previous methods in terms of both performance and stability. Specifically, PointMLP~\citep{ma2022rethinking} obtains $94.1\%$ on ModelNet40, but we surpass it by over $3\%$ in ScanObjectNN and also show more stable performance as indicated by the smaller standard derivation. PointNeXt~\citep{qian2022pointnext} is the current state-of-the-art method, but we still outperform it by a clear margin, especially on mAcc ($87.2\%$ \vs~$85.8\%$). Also, the smallest gap between mean of class-wise accuracy (mAcc) and overall accuracy (OA) achieved by CDFormer implies that our approach shows excellent robust performance on each class instead of biasing to a specific category. We owe it to the proposed collect-and-distribute mechanism for effectively capturing both short- and long-range contexts, which can handle different categories with different scales and shapes.

\subsection{Point Cloud Part Segmentation}
\textbf{Dataset and Metrics.} ShapeNetPart~\citep{yi2016shapnetpart} is a popular dataset for object part segmentation, which is composed of 16,880 3D models from 16 different shape categories (\eg, \textit{``airplane''} and \textit{``chair''}), where 14,006 models for training and 2,874 for testing. For each category, its number of parts is between 2 and 6, and there are total 50 different parts. For evaluation metrics, we report the instance mIoU along with the throughput speed ($instance/second$).

\begin{figure*}[t]
  \centering
  \includegraphics[width=\linewidth]{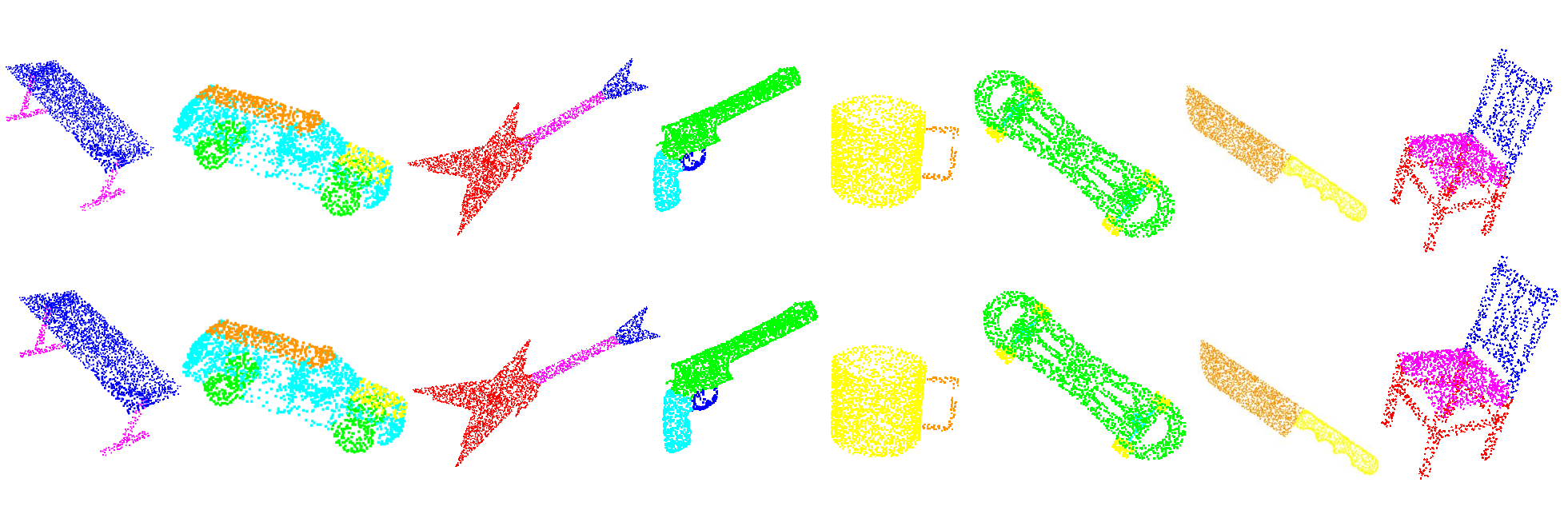}
  \caption{Visualizations of object part segmentation on multiple different categories from the ShapeNetPart dataset. The top row represents ground truth and the predictions from the proposed CDFormer are in the second row.} 
  \label{fig:shapepart}
\end{figure*}

\vspace{1mm}
\noindent
\textbf{Results.} Table~\ref{tab:shapepart} demonstrates the results of CDFormer compared to previous approaches. Our CDFormer outperforms congeneric transformer-based methods, \eg, Point Transformer~\citep{zhao2021point} and Stratified Transformer~\citep{lai2022stratified}, and other representative approaches like KPConv~\citep{thomas2019kpconv}, while it is comparable to the best results of PointNext~\citep{qian2022pointnext} ($87.0\%$ \vs~$87.0\%$). As we will see from Table~\ref{tab:s3dis_cross} that the CDFormer defeats PointNext on S3DIS ($76.0\%$ \vs~$74.9\%$), we thus conjecture that the collect-and-distribute mechanism in CDFormer can not fully exploit the advantage of capturing long-range dependencies on the small size of point cloud considering the huge difference on the size of samples in ShapeNetPart (2,048) and S3DIS (up to 80,000 during training).
The part segmentation results of multiple objects are visualized in Figure~\ref{fig:shapepart}. As observed, the predictions from CDFormer are semantically reasonable and close to the ground truth, further validating its effectiveness.

\begin{table}[t]
    \centering
    \caption{Results on ShapeNetPart.}
    \begin{tabular}{lcc}
    \toprule
    Method  & mIoU (\%)  & Speed \\ %
    \midrule
    PointNet~\citep{qi2017pointnet} & 83.7 & \textbf{1184}\\
    PointNet++~\citep{qi2017pointnet++} & 85.1 & \underline{708} \\
    DGCNN~\citep{wang2019dynamic}  & 85.2 & 147\\
    ASSANet-L~\citep{qian2021assanet} & 86.1 & 640 \\
    PointMLP~\citep{ma2022rethinking} & 86.1 & 270\\
    KPConv~\citep{thomas2019kpconv}& 86.4 & 44 \\
    PT~\citep{zhao2021point} & 86.6 & 297 \\ %
    StratifiedFormer~\citep{lai2022stratified} & 86.6 &  398 \\ %
    CurveNet~\citep{xiang2021walk}   & \underline{86.8} & 97 \\
    PointNeXt~\citep{qian2022pointnext} &\textbf{87.0} & 76\\ %
    \midrule
    CDFormer \small{(ours)} & \textbf{87.0} & 84 \\   %
    \bottomrule
    \end{tabular}
    \label{tab:shapepart}
\end{table}

\begin{table*}[ht]
\centering
\caption{Results on S3DIS using 6-fold cross-validation. The \textbf{bold} and \underline{underline} denote the first and second best performances. Note that $^{*}$ indicates additional 2D images are used and $^{\dagger}$ represents the reproduced results.}
\resizebox{\linewidth}{!}{%
\begin{tabular}{l c c c  c c c c c c c c c c c c c}
\toprule
 Method & mIoU & mACC & OA & \rotatebox{90}{ceiling} & \rotatebox{90}{floor} & \rotatebox{90}{wall} & \rotatebox{90}{beam} & \rotatebox{90}{column} & \rotatebox{90}{window} & \rotatebox{90}{door} & \rotatebox{90}{table} & \rotatebox{90}{chair} & \rotatebox{90}{sofa} & \rotatebox{90}{bookcase} & \rotatebox{90}{board} & \rotatebox{90}{clutter} \\
\midrule
PointNet~\citep{qi2017pointnet}    & 47.6 & 66.2 & 78.6 & 88.0 & 88.7 & 69.3 & 42.4 & 23.1 & 47.5 & 51.6 & 54.1 & 42.0 & 9.6 & 38.2 & 29.4 & 35.2 \\
RSNet~\citep{huang2018recurrent}   & 56.5 & 66.5 & -  & 92.5 & 92.8 & 78.6 & 32.8 & 34.4 & 51.6 & 68.1 & 59.7 & 60.1 & 16.4 & 50.2 & 44.9 & 52.0 \\
PointCNN~\citep{li2018pointcnn}    & 65.4 & 75.6 & 88.1 & {94.8} & {97.3} & 75.8 & 63.3 & 51.7 & 58.4 & 57.2 & 71.6 & 69.1 & 39.1 & 61.2 & 52.2 & 58.6 \\
PointWeb~\citep{zhao2019pointweb}  & 66.7 & 76.2 & 87.3 & 93.5 & 94.2 & 80.8 & 52.4 & 41.3 & 64.9 & 68.1 & 71.4 & 67.1 & 50.3 & 62.7 & 62.2 & 58.5 \\
ShellNet~\citep{Zhang2019ShellNetEP} & 66.8 & - & 87.1    & 90.2 & 93.6 & 79.9 & 60.4 & 44.1 & 64.9 & 52.9 & 71.6 & {84.7} & 53.8 & 64.6 & 48.6 & 59.4 \\
RandLA-Net~\citep{hu2020randla}    & 70.0 & 82.0 & 88.0 & 93.1 & 96.1 & 80.6 & 62.4 & 48.0 & 64.4 & 69.4 & 69.4 & 76.4 & 60.0 & 64.2 & 65.9 & 60.1 \\
KPConv~\citep{thomas2019kpconv}    & 70.6 & 79.1 & -  & 93.6 & 92.4 & 83.1 & 63.9 & 54.3 & 66.1 & 76.6 & 57.8 & 64.0 & 69.3 & {74.9} & 61.3 & 60.3 \\
SCF-Net~\citep{fan2021scf}         & 71.6 & 82.7 & 88.4 & 93.3 & 96.4 & 80.9 & 64.9 & 47.4 & 64.5 & 70.1 & 71.4 & 81.6 & 67.2 & 64.4 & 67.5 & 60.9 \\
BAAF~\citep{qiu2021semantic}       & 72.2 & \underline{83.1} & 88.9 & 93.3 & 96.8 & 81.6 & 61.9 & 49.5 & 65.4 & 73.3 & 72.0 & {83.7} & 67.5 & 64.3 & 67.0 & 62.4 \\
CBL~\citep{tang2022contrastive}    & 73.1 &  79.4 & 89.6 & 94.1 & 94.2 & {85.5} & 50.4 &  58.8 & {70.3} & 78.3 &  75.7 & 75.0 &  71.8 & {74.0} & 60.0 &  62.4 \\
PT~\citep{zhao2021point} & 73.5 & 81.9 & 90.2 & - & - & - & - & - & - & - & - & - & - & - & - & - \\
DeepViewAgg~\citep{robert2022learning}$^{*}$ & 74.7 & - & - & 90.0 & 96.1 & 85.1 & {66.9} & 56.3 & {71.9} & {78.9} & {79.7} & 73.9 & 69.4 & 61.1 & {75.0} & {65.9} \\
PointNext-XL~\citep{qian2022pointnext} & 74.9 & 83.0 & 90.3 & - & - & - & - & - & - & - & - & - & - & - & - & - \\
PointNext-XL~\citep{qian2022pointnext}$^{\dagger}$ & \underline{74.9} & 83.0 & \underline{90.3} & 94.1 & 96.8 & 85.0 & 61.4 & {64.2} & 68.5 & {78.7} & 76.9 & 70.2 & {74.3} & 70.7 & 69.9 & 63.2 \\
\midrule
CDFormer & \textbf{76.0} & \textbf{84.6} & \textbf{90.7} & {94.4} & {97.8} & {86.7} & {70.8} & {66.7} & 69.1 & {78.9} & {77.8} & 64.9 & {75.3} & 71.4 & {71.1} & {63.6} \\
\bottomrule
\end{tabular}
}
\label{tab:s3dis_cross}
\end{table*}

\subsection{Point Cloud Scene Segmentation}
\noindent
\textbf{Dataset and Metrics.} S3DIS~\citep{armeni20163d} is a widely used benchmark for point cloud scene segmentation, containing 271 rooms in 6 areas collected from three buildings. The point is annotated with 13 semantic categories such as \textit{``ceiling''} and \textit{``bookcase''}. Following~\cite{qi2017pointnet++,zhao2021point}, we evaluate the proposed method on the Area 5 and also perform the standard 6-fold cross-validation. ScanNetV2~\citep{dai2017scannet} is another challenging dataset that consists of 1,513 room scenes. Among these scenes, 1,201 are used for training and 312 for validation purposes. Each sampled point is assigned a semantic label from one of the available 20 categories, including \textit{``floor"} and \textit{``table"}. Similar to previous methods~\citep{han2020occuseg,liu2021one,wu2022point,yang2023swin3d}, we make evaluations on the validation and testing sets. Regarding the metrics, we report performance using mean class-wise intersection over union (mIoU), mean of class-wise accuracy (mAcc), and overall point-wise accuracy (OA) as in ~\cite{qian2022pointnext,wu2022point}.

\begin{figure*}[p!]
  \centering
  \includegraphics[width=\linewidth]{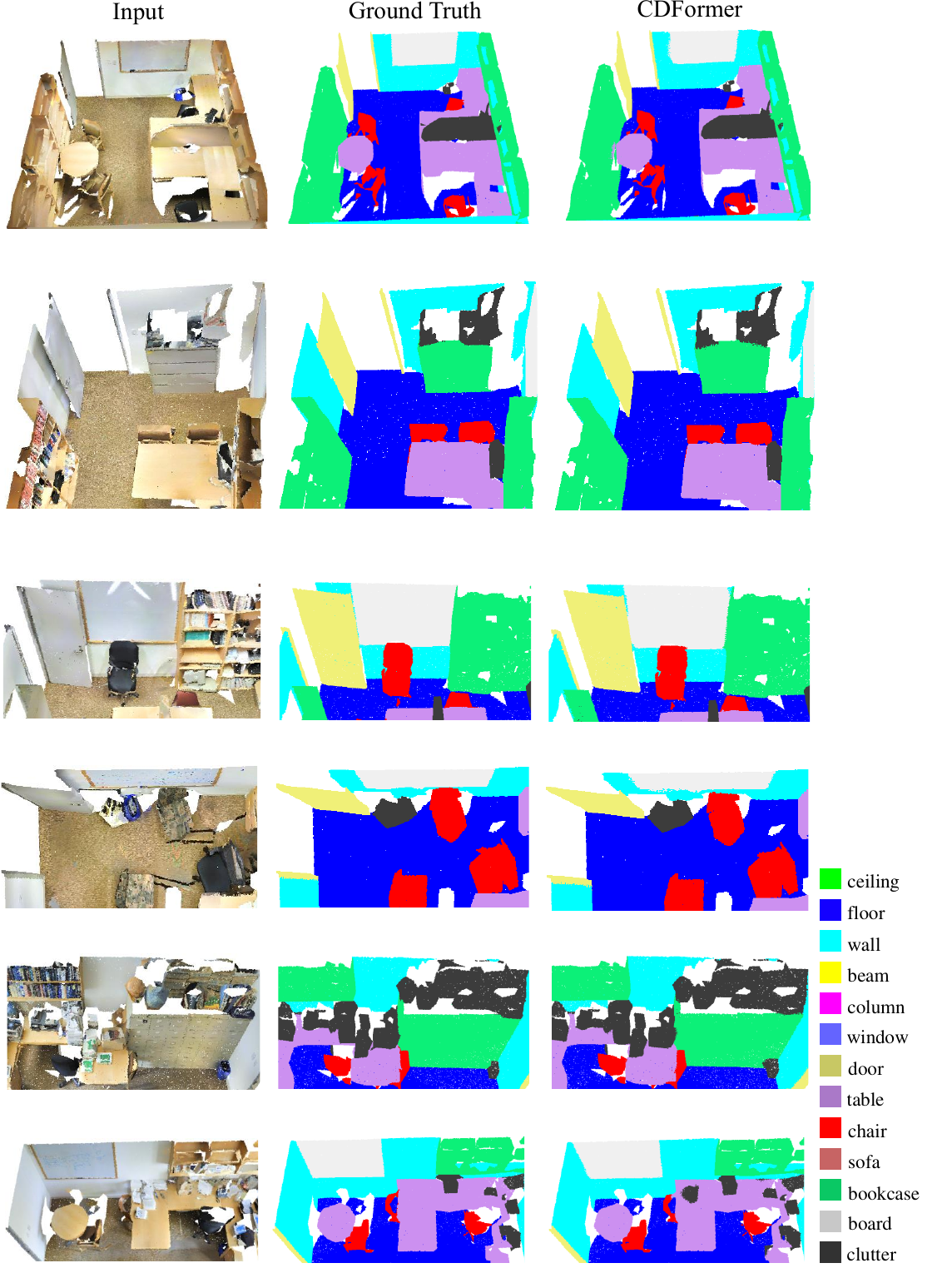}
  \caption{Visualizations of semantic predictions on Area 5 of S3DIS by comparing CDFormer with the ground truth.} 
  \label{fig:s3dis_supp_3}
\end{figure*}

\vspace{1mm}
\noindent\textbf{Results.}
In Table~\ref{tab:s3dis_cross} and~\ref{tab:s3dis_area5}, the proposed CDFormer achieves new state-of-the-art performances on both Area 5 and standard 6-fold cross-validation. Notably, the recent approach PointNext~\citep{qian2022pointnext} has 41.6M parameters for its best results, while our CDFormer with 25.7M parameters achieves a clear improvement of $1.7\%$ and 1.1\% mIoU under Area 5 and 6-fold cross-validation. Additionally, with and without using 2D images, DeepViewAgg~\citep{robert2022learning} obtains $69.5\%$ and $74.7\%$ respectively, which nevertheless is also inferior to CDFormer ($76.0\%$). Additionally, we provide visualizations in Figure~\ref{fig:s3dis_supp_3}. The proposed CDFormer exhibit exceptional ability in capturing short- and long-range contexts, and accurately segmenting complicated scenes. We also present the results on ScanNetV2 in Table~\ref{tab:scannet}. As observed, our CDFormer achieves the best performance of 76.2\% on the validation set and 76.6\% on the testing set, further demonstrating its superiority.

\begin{table*}[ht]
\centering
\caption{Results of the proposed CDFormer and recent state-of-the-arts on Area 5 of S3DIS. The \textbf{bold} and \underline{underline} denote the first and second best performances.}
\resizebox{\linewidth}{!}{%
\begin{tabular}{l c c c  c c c c c c c c c c c c c}
\toprule
  Method & mIoU & mACC & OA & \rotatebox{90}{ceiling} & \rotatebox{90}{floor} & \rotatebox{90}{wall} & \rotatebox{90}{beam} & \rotatebox{90}{column} & \rotatebox{90}{window} & \rotatebox{90}{door} & \rotatebox{90}{table} & \rotatebox{90}{chair} & \rotatebox{90}{sofa} & \rotatebox{90}{bookcase} & \rotatebox{90}{board} & \rotatebox{90}{clutter} \\
\midrule
PointNet   \citep{qi2017pointnet}              & 41.1   & 49.0 & -   & 88.8 & 97.3 & 69.8 & 0.1 & 3.9  & 46.3 & 10.8 & 59.0 & 52.6 & 5.9  & 40.3 & 26.4 & 33.2  \\
SegCloud   \citep{tchapmi2017segcloud}       & 48.9    & 57.4 & -  & 90.1 & 96.1 & 69.9 & 0.0 & 18.4 & 38.4 & 23.1 & 70.4 & 75.9 & 40.9 & 58.4 & 13.0 & 41.6  \\
PointCNN   \citep{li2018pointcnn}              & 57.3  & 63.9 & 85.9 & 92.3 & 98.2 & 79.4 & 0.0 & 17.6 & 22.8 & 62.1 & 74.4 & 80.6 & 31.7 & 66.7 & 62.1 & 56.7  \\
PCT\citep{guo2021pct}                           & 61.3 & 67.7 & - & 92.5 & 98.4 & 80.6 & 0.0 & 19.4 & 61.6 & 48.0 & 76.6 & 85.2 & 46.2 & 67.7 & 67.9 & 52.3 \\
HPEIN \citep{jiang2019hierarchical}             & 61.9 & 68.3 & 87.2 & 91.5 & 98.2 & 81.4 & 0.0 & 23.3 & 65.3 & 40.0 & 75.5 & 87.7 & 58.5 & 67.8 & 65.6 & 49.4  \\
MinkowskiNet \citep{choy20194d}           & 65.4 & 71.7 & -    & 91.8 & 98.7 & 86.2 & 0.0 & 34.1 & 48.9 & 62.4 & 81.6 & 89.8 & 47.2 & 74.9 & 74.4 & 58.6  \\
KPConv  \citep{thomas2019kpconv}            & 67.1  & 72.8 & -   & 92.8 & 97.3 & 82.4 & 0.0 & 23.9 & 58.0 & 69.0 & 81.5 & 91.0 & 75.4 & 75.3 & 66.7 & 58.9  \\
CGA-Net\citep{lu2021cga}       & 68.6 & - & - & 94.5 & 98.3 & 83.0 & 0.0 & 25.3 & 59.6 & 71.0 & 92.2 & 82.6 & 76.4 & 77.7 & 69.5 & 61.5 \\
CBL~\citep{tang2022contrastive} & 69.4 & 75.2 & 90.6 & 93.9 & 98.4 & 84.2 & 0.0 & 37.0 & 57.7 & 71.9 & 91.7 & 81.8 & 77.8 & 75.6 & 69.1 & 62.9 \\
PointFormer~\citep{zhao2021point} & 70.4 & 76.5 & 90.8 & 94.0 & 98.5 & 86.3 & 0.0 & 38.0 & 63.4 & 74.3 & 89.1 & 82.4 & 74.3 & 80.2 & 76.0 & 59.3 \\
PointNext-XL~\citep{qian2022pointnext}  & 70.5 & 76.8 & 90.6  &  94.2  & 98.5  & 84.4  & 0.0  & 37.7  & 59.3  & 74.0  & 83.1  & 91.6  & 77.4  & 77.2  & 78.8  & 60.6 \\
StratifiedFormer~\citep{lai2022stratified} & \underline{72.0}  & \underline{78.1} & \textbf{91.5}  & 96.2 & 98.7 & 85.6 &  0.0 &  46.1 &  60.0 &  76.8 &  92.6 &  84.5 &  77.8 &  75.2 &  78.1 &  64.0 \\ %
\midrule
CDFormer & \textbf{72.2} & \textbf{78.5} & \underline{91.2} & 95.1 & 98.8 & 86.3 & 0.0 & 49.3 & 62.4 & 72.1 & 83.5 & 92.3 & 83.9 & 76.1 & 75.9 & 62.7 \\
\bottomrule
\end{tabular}
}%
\label{tab:s3dis_area5}
\end{table*}

\begin{table*}[ht]
\centering
\caption{Results of mIoU (\%) on both validation and testing set (including per-category IoU) of ScanNetV2. }
\resizebox{\linewidth}{!}{%
\begin{tabular}{l c c c c c c c c c c c}
\toprule
  \rotatebox{0}{Method} & \rotatebox{0}{Val} & \rotatebox{0}{Test} &
  \rotatebox{0}{bathtub}&	\rotatebox{0}{bed}&	\rotatebox{0}{bookshelf}	&\rotatebox{0}{cabinet}	&\rotatebox{0}{chair}	&\rotatebox{0}{counter}	&\rotatebox{0}{curtain}	&\rotatebox{0}{desk}	&\rotatebox{0}{door}	\\
\midrule
PointNet++~\citep{qi2017pointnet++} & 53.5 & 55.7 & 73.5 & 66.1 & 68.6 & 49.1 & 74.4 & 39.2 & 53.9 & 45.1 & 37.5   \\
RandLA-Net~\citep{hu2020randla} & - & 64.5 & 77.8 & 73.1 & 69.9 & 57.7 & 82.9 & 44.6 & 73.6 & 47.7 & 52.3  \\
PointConv~\citep{wu2019pointconv}     & 61.0    &  66.6  &  78.1  &  75.9  &  69.9  &  64.4  &  82.2  &  47.5  &  77.9  &  56.4  &  50.4   \\
PointASNL~\citep{yan2020pointasnl}    & 63.5    &  66.6  &  70.3  &  78.1  &  75.1  &  65.5  &  83.0  &  47.1  &  76.9  &  47.4  &  53.7    \\
KPConv~\citep{thomas2019kpconv}     & 69.2  &  68.4  &  84.7  &  75.8  &  78.4  &  64.7  &  81.4  &  47.3  &  77.2  &  60.5  &  59.4    \\
CBL~\citep{tang2022contrastive} & - & 70.5 & 76.9 & 77.5 & 80.9 & 68.7 & 82.0 & 43.9 & 81.2 & 66.1 & 59.1  \\
PointNext~\citep{qian2022pointnext}  & 71.5 & 71.2 & - & - & - & - & - & - & - & - & -   \\
PointMeta~\citep{lin2022meta} & 71.4 & 83.5 & 78.5 & 82.1 & 68.4 & 84.6 & 53.1 & 86.5 & 61.4 & 59.6   \\
SparseCNN~\citep{graham20183d} & 72.8 & 72.5 & 64.7 & 82.1 & 84.6 & 72.1 & 86.9 & 53.3 & 75.4 & 60.3 & 61.4  \\
MinkUNet~\citep{choy20194d}       & 72.2  &  73.6  &  85.9  & {81.8} & {83.2} &  70.9  & {84.0} & {52.1} & {85.3} &  66.0  &  64.3   \\
StratifiedFormer~\citep{lai2022stratified}  & 74.3 & 74.7 & 90.1 & 80.3 & 84.5 & 75.7 & 84.6 & 51.2 & 82.5 & 69.6 & 64.5   \\
BPNet~\citep{hu2021bidirectional}     & 73.9   &  74.9  &  90.9  &  81.8  &  81.1  &  75.2  &  83.9  &  48.5  &  84.2  &  67.3  &  64.4   \\
PTv2~\citep{wu2022point} & 75.4 & 75.2 & 74.2 & 80.9 & 87.2 & 75.8 & 86.0 & 55.2 & 89.1 & 61.0 & 68.7  \\
\rowcolor{lightgray} CDFormer \small{(ours)} & \textbf{76.2}& \textbf{76.6} & 77.9 & 79.2 & 86.1 & 75.3 & 88.4 & 59.1 & 88.7 & 62.6 & 71.3  \\
\midrule
\rotatebox{0}{Method} &\rotatebox{0}{floor} & \rotatebox{0}{other.}	&\rotatebox{0}{picture}	&\rotatebox{0}{refrig.}	&\rotatebox{0}{shower.}	&\rotatebox{0}{sink}	&\rotatebox{0}{sofa}	&\rotatebox{0}{table}	&\rotatebox{0}{toilet}	&\rotatebox{0}{wall}	&\rotatebox{0}{window}\\
\midrule
PointNet++~\citep{qi2017pointnet++} & 94.6 & 37.6 & 20.5 & 40.3 & 35.6 & 55.3 & 64.3 & 49.7 & 82.4 & 75.6 & 51.5  \\
RandLA-Net~\citep{hu2020randla} & 94.5 & 45.4 & 26.9 & 48.4 & 74.9 & 61.8 & 73.8 & 59.9 & 82.7 & 79.2 & 62.1 \\
PointConv~\citep{wu2019pointconv}    &  95.3     &   42.8  &  20.3  &  58.6  &  75.4  &  66.1  &  75.3  & 58.8  & {90.2} &  81.3  &  64.2  \\
PointASNL~\citep{yan2020pointasnl}    &  95.1   &  47.5  &  27.9  &  63.5  &  69.8  &  67.5  &  75.1  & 55.3  &  81.6  &  80.6  &  70.3  \\
KPConv~\citep{thomas2019kpconv}     &  93.5  &   45.0  &  18.1  &  58.7  &  80.5  &  69.0  &  78.5  & 61.4  &  88.2  &  81.9  &  63.2  \\
CBL~\citep{tang2022contrastive} & 94.5 & 51.5 & 17.1 & 63.3 & 85.6 & 72.0 & 79.6 & 66.8 & 88.9 & 84.7 & 68.9 \\
PointNext~\citep{qian2022pointnext} & - &  - & - & - & - & - & - & - & - & - & -   \\
PointMeta~\citep{lin2022meta} & 95.3 & 50.0 & 24.6 & 67.4 & 88.8 & 69.2 & 76.4 & 62.4 & 84.9 & 84.4 & 67.5 \\
SparseCNN~\citep{graham20183d} & 95.5 & 57.2 & 32.5 & 71.0 & 87.0 & 72.4 & 82.3 & 62.8 & 93.4 & 86.5 & 68.3 \\
MinkUNet~\citep{choy20194d}  &  95.1 & {54.4} &  28.6  &  73.1  & {89.3} &  67.5  &  77.2  & 68.3  &  87.4  &  85.2  & {72.7} \\
StratifiedFormer~\citep{lai2022stratified} & 95.6 & 57.6 & 26.2 & 74.4 & 86.1 & 74.2 & 77.0 & 70.5 & 89.9 & 86.0 & 73.4 \\
BPNet~\citep{hu2021bidirectional}     &  95.7  &  52.8  &  30.5  &  77.3  &  85.9  &  78.8  &  81.8  & 69.3  &  91.6  &  85.6  &  72.3  \\
PTv2~\citep{wu2022point}  & 96.0 & 55.9 & 30.4 & 76.6 & 92.6 & 76.7 & 79.7 & 64.4 & 94.2 & 87.6 & 72.2 \\
\rowcolor{lightgray} CDFormer \small{(ours)} & 97.6 & 57.7 & 33.4 & 69.8 & 95.1 & 82.0 & 82.8 & 68.3 & 95.4 & 89.3 & 72.2 \\
\bottomrule
\end{tabular}
}%
\label{tab:scannet}
\end{table*}

\subsection{Ablation Studies}
We perform all the ablation studies on Area 5 of S3DIS.
\begin{table}[t]
    \centering
    \caption{Ablation study on the CD mechanism. 
    }
    \resizebox{\linewidth}{!}{
    \begin{tabular}{p{0.4cm}  p{0.5cm}<{\centering} p{0.8cm}<{\centering} p{1.2cm}<{\centering}  p{1.3cm}<{\centering} p{1.4cm}<{\centering} p{1.2cm}<{\centering}}
    \toprule
    cfg & LSA & Collect & Distribute & mIoU(\%) & mAcc (\%) & OA (\%) \\ 
    \midrule
    \romannumeral1 & $\checkmark$ &              &              & 67.6 & 74.1 & 89.7 \\
    \romannumeral2 & $\checkmark$ & $\checkmark$ &              & 70.1 & 76.6 & 90.7  \\
    \romannumeral3 & $\checkmark$ &              & $\checkmark$ & 70.3 & 77.0 & 90.7 \\ 
    \romannumeral4 & $\checkmark$ & $\checkmark$ & $\checkmark$ & \textbf{72.2} & \textbf{78.5} & \textbf{91.2} \\
    \bottomrule
    \end{tabular}}
    \label{tab:attention}
\end{table}
\vspace{1mm}

\noindent\textbf{Collect-and-Distribute Mechanism.} To explore the effectiveness of the proposed collect-and-distribute (CD) mechanism, we use the counterpart vanilla parameter-free (or tiny) operations as baselines. Specifically, we only keep the max operation to aggregate local features while discarding neighbor self-attention when no \textit{collect}. Also, we adopt an interpolation layer followed by a simple MLP to add back to local points if no \textit{distribute}. In Table~\ref{tab:attention}, we find that without the collect-and-distribute mechanism, it achieves a relatively low mIoU $67.6\%$, which is then improved by \textit{collect} for catching the long-range dependencies to $70.1\%$. Furthermore, when further distributing the long-range contexts back to the local points, we achieve the best performances on all metrics, \ie, $72.2\%$ mIoU, $78.5\%$ mAcc, and $91.2\%$ OA. Besides, if without \textit{collect} to explicitly and broadly explore long-range dependencies, using the max aggregation with \textit{Distribute} (ses \romannumeral3) significantly drops the performance from $72.2\%$ to $70.3\%$ on mIoU.

\begin{table}[t]
    \centering
    \caption{Ablation study on CAPE.
     }
    \resizebox{\linewidth}{!}{
    \begin{tabular}{p{0.5cm}  p{0.5cm}<{\centering} p{0.5cm}<{\centering} p{0.5cm}<{\centering}  p{1.5cm}<{\centering} p{1.5cm}<{\centering} p{1.2cm}<{\centering}}
    \toprule
    cfg & QKV & CA & Plain & mIoU (\%) & mAcc (\%) & OA (\%) \\ 
    \midrule
    \romannumeral1 &              &              &              & 68.9 & 76.6 & 89.7 \\
    \romannumeral2 & $\checkmark$ &              &              & 69.7 & 76.3 & 91.0 \\
    \romannumeral3 & $\checkmark$ & $\checkmark$ &              & \textbf{72.2} & \textbf{78.5} & \textbf{91.2} \\
    \romannumeral4 & $\checkmark$ &              & $\checkmark$ & 70.8 & 77.5 & 90.9 \\ 
    \bottomrule
    \end{tabular}}
    \label{tab:position}
\end{table}
\vspace{1mm}
\noindent\textbf{Context-Aware Position Encoding.} To evaluate the influence of position encoding,  we compare the proposed method with and without using the position encoding in the query, key, and value (\textbf{QKV}), respectively. We also compare the context-aware (\textbf{CA}) format of position encoding with a vanilla one (\ie, \textbf{Plain}) as in~\cite{dosovitskiy2020image}. 
As shown in Table~\ref{tab:position}, without the position encoding, we have dissatisfied results that are consistent with the observations in ~\cite{lai2022stratified,zhao2021point}. Next, we employ a normal encoding without context-aware format (see \romannumeral2) on QKV, \ie, no interaction with the contexts Q, K. This time, it does work but with only marginal improvement. Lastly, when using the proposed CAPE (see \romannumeral3), it can achieve the state-of-the-art performance $72.2\%$. Through the dynamic adaptation based on input contexts, the position clues are further enhanced in the proposed CDFormer to effectively learn local-global relations. Additionally, we also show the results of using a vanilla version of position encoding~\citep{dosovitskiy2020image} (see \romannumeral4) for a fair comparison. 

\begin{figure*}[t]
  \centering
  \includegraphics[width=\linewidth]{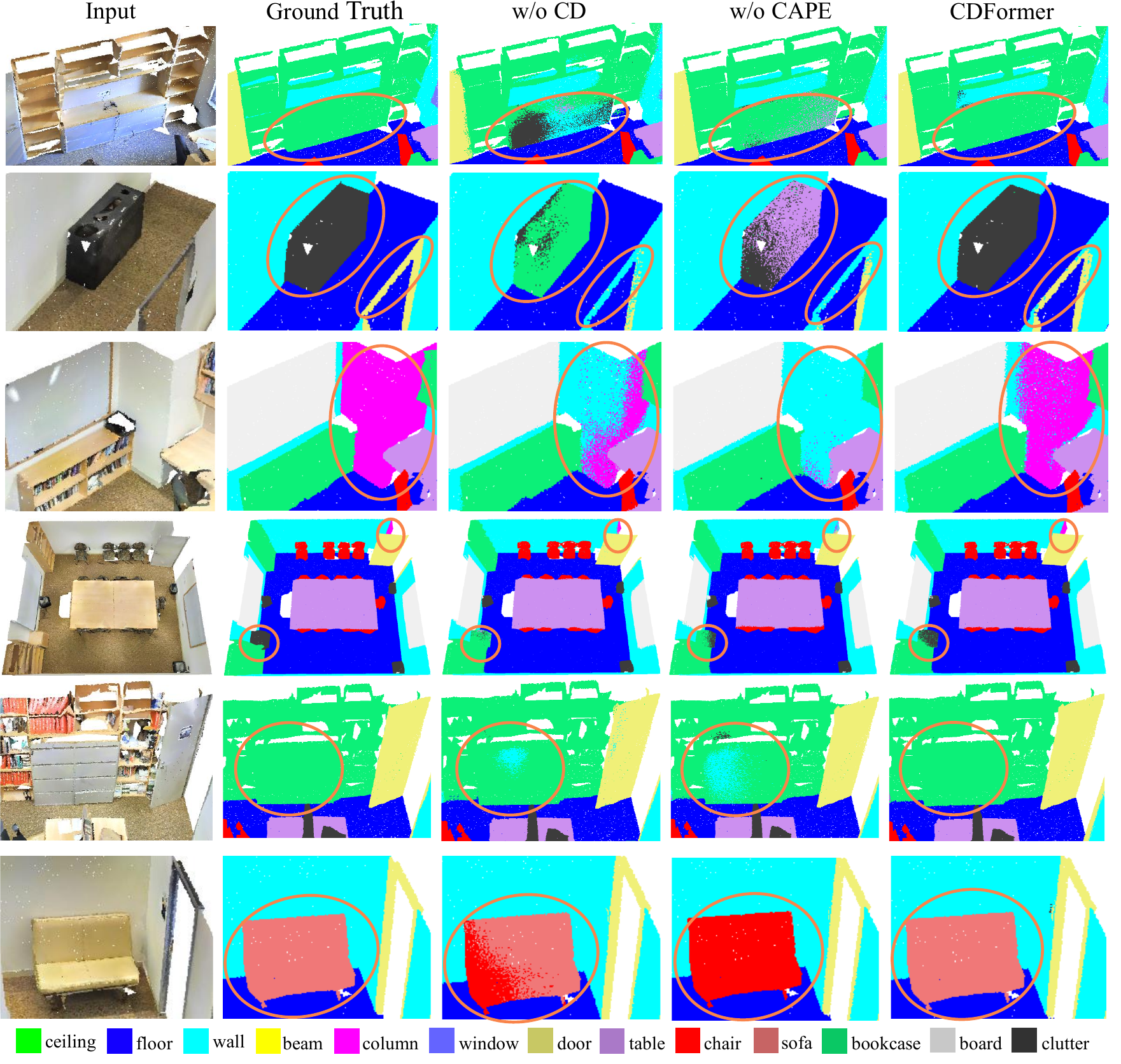}
  \caption{Visualizations of scene semantic segmentation on Area 5 of S3DIS by comparing CDFormer with its baselines and ground truth.
  The improved areas against baselines are highlighted by the {\color{orange} orange} circles.} 
  \label{fig:s3dis_supp_compare}
\end{figure*}

\vspace{1mm}
\noindent\textbf{Visualization Comparisons.}
To further demonstrate the improvement of CDFormer against its baselines, we provide visualizations in Figure~\ref{fig:s3dis_supp_compare} by comparing to \textbf{w/o CD} (see {\romannumeral1} in Table~\ref{tab:attention}) and \textbf{w/o CAPE} (see {\romannumeral2} in Table~\ref{tab:position}). Taking the first row as an example, without the collect-and-distribute mechanism, the model fails to utilize the semantic-correct long-range contexts and thus can not precisely segment local points; without CAPE, it loses the enhanced position clues for better communication within points and thus obtains dissatisfied results; the CDFormer achieves the convincing predictions as expected.

\vspace{1mm}
\noindent\textbf{Model Scalability.}
We evaluate the model scalability under four stages with $[2, 2, 6, 2]$ blocks via the following three configurations:
\vspace{-1mm}
\begin{itemize}[noitemsep,leftmargin=1.5em]
    \item CDFormer-S: $\mathcal{C}=[16,32,64,128], \mathcal{H}=[1,2,4,8]$
    \item CDFormer-B: $\mathcal{C}=[32,64,128,256], \mathcal{H}=[2,4,8,16]$
    \item CDFormer-L: $\mathcal{C}=[48,96,192,384], \mathcal{H}=[3,6,12,24]$
\end{itemize}
\vspace{-1mm}
As shown in Table~\ref{tab:scale}, there are clear performance improvements when using a larger model.For example, the improvements are very impressive in terms of both mIoU and mAcc, \ie, mIoU: $67.6\% \rightarrow 69.7\% \rightarrow 72.2\%$; mAcc: $74.9\% \rightarrow 77.1\% \rightarrow 78.5\%$. Notably, the consistent improvements by increasing the model size illustrate the extraordinary scalability of the proposed CDFormer for point cloud analysis.

\begin{table}[t]
    \centering
    \caption{Ablation study on model scalability.}
    \resizebox{\linewidth}{!}{
    \begin{tabular}{lrccc}
    \toprule
    Model & \#Params & mIoU (\%) & mAcc (\%) & OA (\%) \\
    \midrule
    CDFormer-S & 3.1M & 67.6 & 74.9 & 89.8 \\
    CDFormer-B & 11.7M & 69.7 & 77.1 & 90.4 \\
    CDFormer-L & 25.7M & \textbf{72.2} & \textbf{78.5} & \textbf{91.2}  \\
    \bottomrule
    \end{tabular}}
    \label{tab:scale}
\end{table}
\begin{table}[t]
    \centering
    \caption{Ablation study on number of neighbors.}
    \resizebox{0.8\linewidth}{!}{
    \begin{tabular}{lcccc}
    \toprule
    cfg & $K$ & mIoU (\%) & mAcc (\%) & OA (\%) \\ 
    \midrule
    \romannumeral1 & 4 & 69.2 & 75.7 & 90.3 \\
    \romannumeral2 & 8 & 71.1 & 77.9 & \textbf{91.4} \\
    \romannumeral3 & 16 & \textbf{72.2} & \textbf{78.5} & 91.2 \\
    \bottomrule
    \end{tabular}}
    \label{tab:knn}
\end{table}
\vspace{1mm}
\noindent\textbf{Number of Neighbors.}
We show the influences for using the different number of neighbors in CD Blocks for patch division, NSA, and NCA. For simplicity, we use the same $K$ for all modules and stages. As show in Table~\ref{tab:knn}, when increasing the number of neighbors from $K=4$ to $K=8$, the improvement ($69.2\% \rightarrow 71.1\%$) is significant. This is possibly because a relatively large number of neighbors can bring better long-range contexts, which help the proposed CDFormer learn the local-global structures. Notably, we also find that the advancement from $K=8$ to $K=16$ is modest. This kind of saturated phenomenon is expected since too much contexts may introduce different categories of semantics that degrade the local features. Similar observations have also been reported in~\cite{zhao2021point}. Therefore, we use $K=16$ as default.

\vspace{1mm}
\noindent
\textbf{Point Coverage.}
In~\Secref{sec:pl}, we employ the patch division on the point cloud to generate $M$ patches with $K$ points in each patch. Specifically, we first utilize the FPS~\citep{eldar1997farthest,qi2017pointnet++} to downsampling the point features $\mX \in \mathbb{R}^{N \times C}$ to local patch centers $\bar{\mX} \in \mathbb{R}^{M \times C}$ where $M=N/S$ and $S$ is the scale factor. After that, we group the $K$ nearest neighbors around each patch center and reformulate $M$ patches as $\hat{\mX} \in \mathbb{R}^{M \times K \times C}$. However, this process may result in some points being left uncovered while others are covered multiple times. Here we present a quantitative analysis. We show the statistics of points covered multiple times or uncovered at different stages of CDFormer on the S3DIS dataset in Table~\ref{tab:point_cover}. As we can see, no more than 0.4\% of all points are uncovered across all stages. In addition, each point is covered by two times on average due to $M = N / 8$ and $MK = 2N$, and maximally covered by six times. As a result, their impact should be negligible.
\begin{table}[ht]
\centering
\caption{Statistics of point coverage on different stages.}
\label{tab:point_cover}
\begin{tabular}{ccc}
\toprule
Stage & Uncovered Ratio & Maximum Times Covered \\
\midrule
1 & 0.28\% & 6.00 \\
2 & 0.28\% & 5.26 \\
3 & 0.38\% & 4.62 \\
4 & 0.24\% & 4.18 \\
\bottomrule
\end{tabular}
\end{table}

\begin{figure*}[h]
  \centering
  \includegraphics[width=\linewidth]{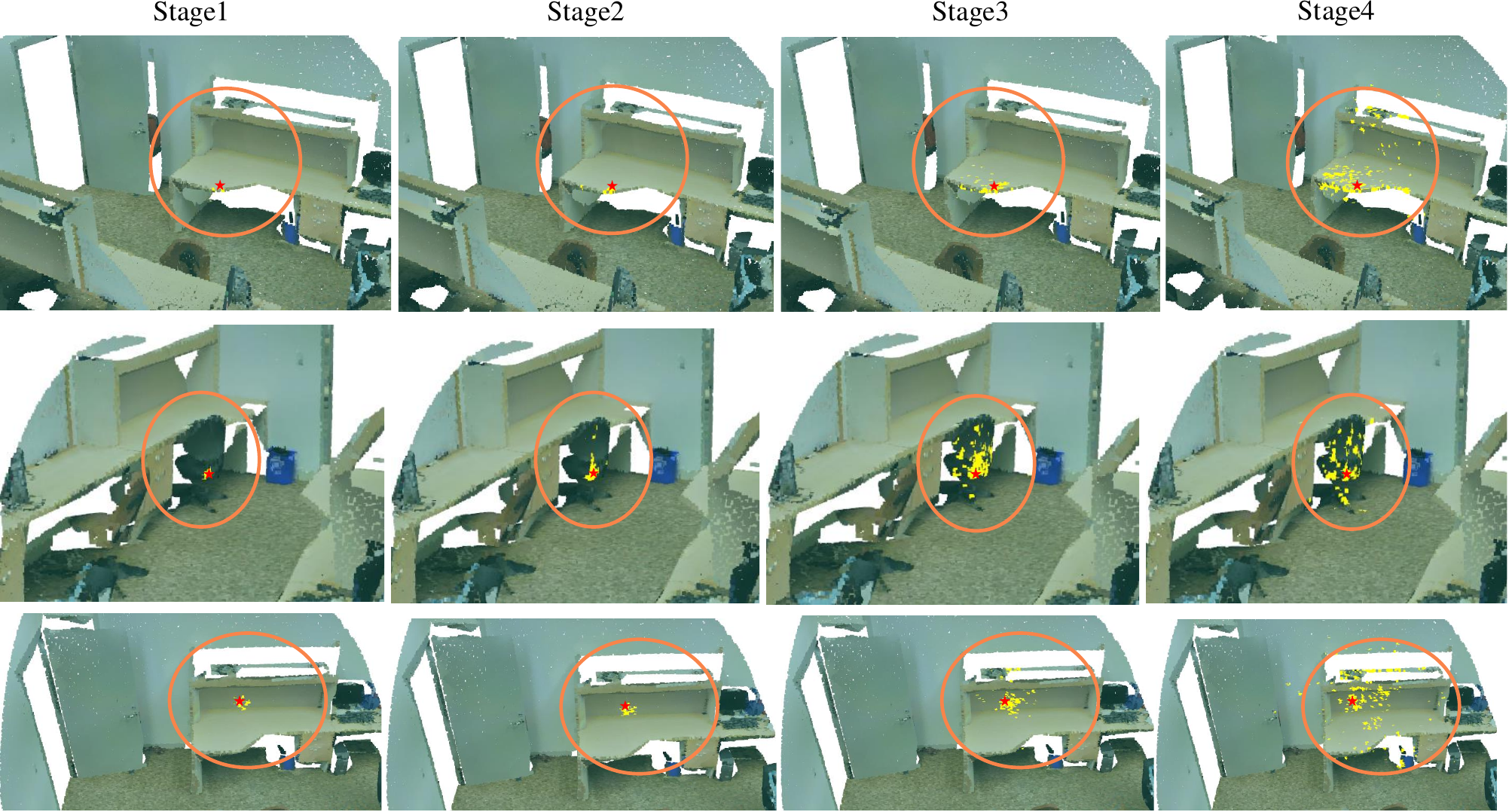}
  \caption{Visualization of activation maps where region of interest are highlighted by {\color{orange} orange} circles at different stages of CDFormer given the feature of interest (marked as {\color{red} red} star).} 
  \label{fig:attn}
\end{figure*}

\vspace{1mm}
\noindent
\textbf{Activation Maps.}
We illustrate the activation maps of different stages of CDFormer to show the effectiveness of collect-and-distribute mechanism in Figure~\ref{fig:attn}. Specifically, we backtrack the feature of interest from each stage and calculate the gradients for each input point. A larger gradient indicates that the current point contributes more to the feature of interest. In rows one to three, we select points of interest from classes \textit{``table''}, \textit{``chair''}, and \textit{``bookcase''} respectively. As observed, the number of perceptive points increases inside the same category as the stage progresses due to the collect-and-distribute mechanism. For example, in the case of a point on the edge of a table, it can first interact with more points along that edge before reaching points inside the table itself. Similarly, for a point on a chair, it gradually captures all relevant points distributed across its surface while avoiding irrelevant or nearby points from other categories.

\section{Conclusion}
\label{sec:conclusion}
In this paper, we introduce the innovative CDFormer for 3D point cloud analysis. With our proposed collect-and-distribute mechanism, it can capture both short- and long-range contexts simultaneously, effectively learning local-global representations of point clouds. Furthermore, we enhance position clues dynamically via context-aware position encoding to facilitate communication among point features. We conducted comprehensive experiments on S3DIS, ScanNetV2, ShapeNetPart, ModelNet40, and ScanObjectNN for segmentation and classification tasks to demonstrate the superiority of CDFormer and the advantages of each design choice.

\bibliographystyle{spbasic}
\bibliography{main.bib}

\end{document}